\documentclass[10pt,twocolumn,letterpaper]{article}

\usepackage{wacv}
\usepackage{times}
\usepackage{epsfig}
\usepackage{graphicx}
\usepackage{amsmath}
\usepackage{amssymb}
\usepackage{booktabs}

\usepackage{multirow}
\usepackage{dblfloatfix}

\def\wacvPaperID{374} 

\wacvalgorithmstrack   

\wacvfinalcopy 

\ifwacvfinal
\usepackage[breaklinks=true,bookmarks=false]{hyperref}
\else
\usepackage[pagebackref=true,breaklinks=true,colorlinks,bookmarks=false]{hyperref}
\fi

\begin{document}

\title{Dynamic Neural Portraits}

\author{Michail Christos Doukas$^{1,2} \quad$ Stylianos Ploumpis$^{2} \quad$ Stefanos Zafeiriou$^{1,2}$\\
$^1$Imperial College London, UK$\quad^2$Huawei Technologies, London, UK\\
{\tt\small $\{$michail.christos.doukas,  stylianos.ploumpis, stefanos.zafeiriou1$\}$@huawei.com}
}

\maketitle
\thispagestyle{empty}

\begin{abstract}
We present Dynamic Neural Portraits, a novel approach to the problem of full-head reenactment. Our method generates photo-realistic video portraits by explicitly controlling head pose, facial expressions and eye gaze. Our proposed architecture is different from existing methods that rely on GAN-based image-to-image translation networks for transforming renderings of 3D faces into photo-realistic images. Instead, we build our system upon a 2D coordinate-based MLP with controllable dynamics. Our intuition to adopt a 2D-based representation, as opposed to recent 3D NeRF-like systems, stems from the fact that video portraits are captured by monocular stationary cameras, therefore, only a single viewpoint of the scene is available. Primarily, we condition our generative model on expression blendshapes, nonetheless, we show that our system can be successfully driven by audio features as well. Our experiments demonstrate that the proposed method is 270 times faster than recent NeRF-based reenactment methods, with our networks achieving speeds of 24 fps for resolutions up to $1024 \times 1024$, while outperforming prior works in terms of visual quality.
\end{abstract}

\vspace{-1em}

\section{Introduction}

Controllable video portrait synthesis is an interesting research topic that captures the attention of both Computer Graphics and Computer Vision communities. A video portrait is defined as a sequence of frames that depict a single individual performing diverse head movements and facial expressions. The person's entire head is contained within the frame's borders, along with a small part of the upper body, i.e. neck and torso, while the subject usually stands in front of a static background. Recent attempts to video portraits synthesis using neural networks have shown very promising results, based either on Generative Adversarial Networks (GANs)~\cite{ganGoodfellow} or Neural Radiance Fields (NeRFs)~\cite{mildenhall2020nerf}. The applications of such systems are numerous, ranging from video editing and movie dubbing to teleconference, virtual assistance, social media, VR and games. 

A number of learning-based solutions for generating video portraits rely on GAN-based image-to-image translation models, with an encoder-decoder architecture. For instance, Deep Video Portraits (DVP)~\cite{kim2018deep} employ a network that learns a mapping from coloured renderings of 3D faces to realistic portraits. Similarly, Head2Head~\cite{koujan2020head2head} translates images of Projected Normalized Coordinate Codes (PNCCs)~\cite{zhu2016face} into photo-realistic frames, using a video-to-video framework. The renderings of 3D faces that serve as conditional input to generative neural networks depend on expression blendshapes that are obtained after fitting 3D Morphable Models (3DMMs)~\cite{blanz1999morphable, paysan20093d, booth2018large, cao2013facewarehouse, li2017learning} to videos. The fitting step is followed by a physically-based rendering process, which creates the 2D renderings. On the contrary, we propose a multi-layer perceptron (MLP) that conditions synthesis directly on non-spatial data (e.g. expression parameters), and thus does not require renderings of 3D faces.

\vspace{-0.1em}

More recently, NerFACE~\cite{Gafni_2021_CVPR} capitalised on the photo-realism achieved by NeRFs~\cite{mildenhall2020nerf} and produced synthetic video portraits of higher quality, in larger resolutions, such as $512 \times 512$. Nonetheless, due to ray casting and volume sampling, image rendering with NerFACE requires several seconds per frame. Moreover, the authors make the assumption that the tracked head pose parameters coincide with the camera viewpoints of the scene, which causes significant inconsistencies in torso synthesis, as in practice the camera is static and therefore only a single viewpoint exists for the scene. AD-NeRF~\cite{guo2021ad} proposes an audio-driven method for portrait synthesis and solves the camera problem by taking advantage of face segmentation maps~\cite{Lee2020MaskGANTD}. They split the portrait into head, torso and background and devise two separate NeRF models: one for the head, which uses head poses as camera viewpoints, and another one for the torso, that treats head poses as simple inputs to the MLP, while considering camera viewpoints fixed. However, AD-NeRF is even slower during inference, as it evaluates two MLPs.

\vspace{-0.1em}

\begin{figure*}[t!]
\centering
\begin{picture}(100,180)
    \put(-198,20){\includegraphics[scale=0.265]{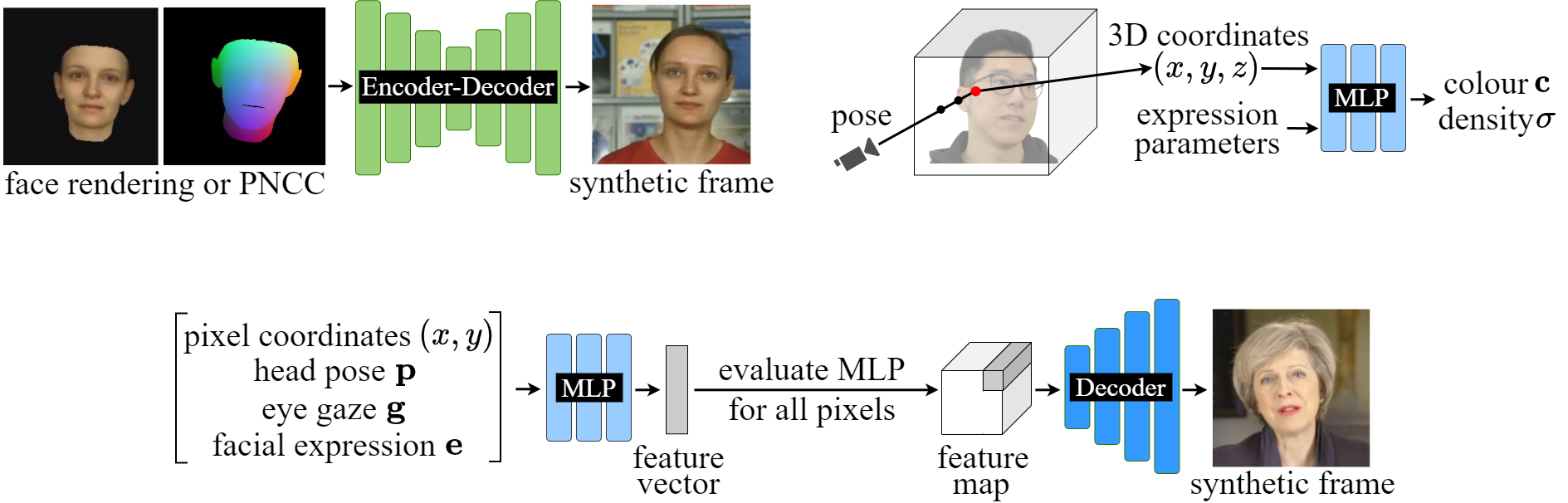}}
    \put(-184,106){(a) Image-based rendering of video portraits \cite{kim2018deep,koujan2020head2head}}
    \put(84,106){(b) NeRF-based video portrait rendering \cite{Gafni_2021_CVPR,guo2021ad}}
    \put(-50,10){(c) Our paradigm for video portrait rendering}
\end{picture}
\caption{In contrast to traditional GAN-based image-to-image translation approaches to full-head reenactment, such as DVP~\cite{kim2018deep} and Head2Head~\cite{koujan2020head2head} (a), or recent NeRF-based video portrait rendering methods, namely NerFACE \cite{Gafni_2021_CVPR} and AD-NeRF \cite{guo2021ad} (b), we propose a novel paradigm for controllable video portrait synthesis, composed of an MLP and a CNN-based decoder (c).}
\label{fig:teaser}
\end{figure*}
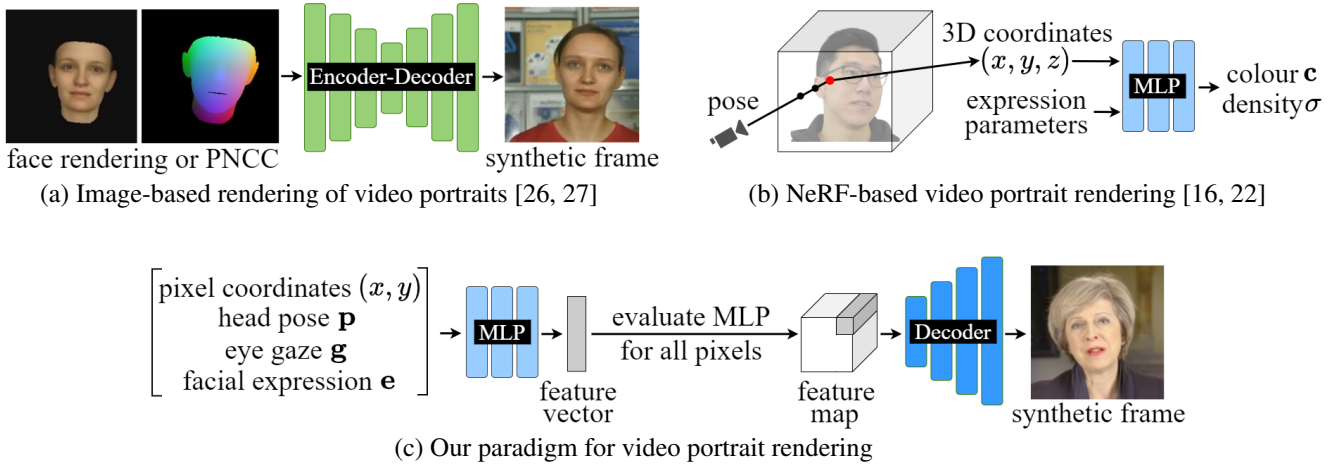

In this paper, we take a completely different approach and present \textit{Dynamic Neural Portraits}, a fast and efficient framework for reenacting human faces. Our method draws inspiration from implicit neural representations (INR), as we leverage neural networks to parameterise video portraits. We follow practices from both conditionally independent pixel synthesis (CIPS) \cite{anokhin2021image} and neural rendering with convolutional networks. To be more specific, we perform video portrait synthesis using a 2D coordinate-based MLP with controllable dynamics. That is, we condition an MLP network on pixel coordinates, expression, pose and gaze parameters, without relying on renderings of 3D faces. Nonetheless, instead of directly predicting pixel colours, we adopt practices from 3D aware GANs~\cite{niemeyer2021giraffe, gu2021stylenerf, zhou2021cips, chan2022efficient} and propose an MLP that produces feature vectors, which are computed across all 2D spatial locations and up-sampled with a CNN-based decoder. We optimise our "hybrid" MLP-CNN architecture jointly on the task of video portrait reconstruction. To the best of our knowledge, we are the first to condition a 2D coordinate-based MLP on expression, pose and gaze parameters for explicitly controlling video portraits, without relying on renderings of 3D faces, GANs, or NeRFs. Moreover, unlike previous methods that focus exclusively either on expression blendshapes or audio signals as driving data, our work demonstrates how to leverage both modalities using the same architecture. Our approach combines high-quality samples with unparalleled execution performance, $270$ times faster than recent NeRF-based state-of-the-art reenactment methods \cite{Gafni_2021_CVPR, guo2021ad}. The contributions of this paper can be summarised as follows:
\begin{itemize}
    \item We propose a new approach to full-head reenactment, with a generator that consists of a 2D coordinate-based MLP with controllable dynamics and a CNN decoder.
    \item We show that our architecture can be driven either by expression blendshapes or audio-based features.
    \item Our comprehensive experiments demonstrate that our method outperforms related state-of-the-art systems, both in terms of execution speed and image quality.
\end{itemize}

\section{Related work}

\noindent{\textbf{3D Face Modeling and Face Reenactment}.} Since their introduction, 3DMMs~\cite{blanz1999morphable, paysan20093d, booth2018large, cao2013facewarehouse, li2017learning} have been widely used to represent human faces, as they are strong statistical models that enable explicit control over the shape and texture of 3D facial meshes. Various 3D face reconstruction methods depend on 3DMMs to recover 3D faces from visual data. Such methods are classified either as optimisation-based~\cite{Vetter98estimatingcoloured,booth20173d}, as they fit 3DMMs to visual data and estimate parameters in an analysis-by-synthesis fashion, or learning-based \cite{ganfit,DECA} that rely on neural networks to reconstruct 3D faces. Apart from capturing coarse meshes, the later have shown very promising results in extracting even local fine details. Recovering facial shape and expression information from video data has been proven very useful for numerous face reenactment methods~\cite{garrido2014automatic,thies2015real,thies2016face}. The work of Garrido \textit{et al.}~\cite{garrido2014automatic} is one of the first attempts to reenact faces while relying on 3D face modeling. A succeeding graphics-based method, namely Face2Face~\cite{thies2016face}, performs real-time expression transfer from a driving sequence of frames to a target identity, by re-writing the interior facial region of the target video. A more recent approach, HeadOn~\cite{thies2018headon}, achieves full-head reenactment that includes pose and gaze transfer, based on RGB-D video data.

\noindent{\textbf{Learning-based Talking Head Synthesis}.} In contrary to the graphics-based systems described above, most of the latest face re-targeting approaches are learning-based \cite{suwajanakorn2017synthesizing,kim2018deep,koujan2020head2head}. Suwajanakorn \textit{et al.}~\cite{suwajanakorn2017synthesizing} are among the first to propose an audio-driven neural network for mapping acoustic signals to photo-realistic frames with accurate lip motions. Following up, Neural Voice Puppetry~\cite{thies2020nvp} employs a module for translating audio features into expression blendshapes, as an intermediate representation. Regarding video-driven methods, Deep Video Portraits (DVP)~\cite{kim2018deep} is one of the earliest GAN-based approaches to full-head reenactment. It relies on an image-to-image translation network that receives as inputs synthetic face renderings of a parametric 3D face model and generates images of the target subject. In a similar direction, Head2Head~\cite{koujan2020head2head,doukas2021head2head++} translates PNCCs~\cite{zhu2016face} into photo-realistic frames, with the help of a video-based GAN for better temporal stability. Deferred Neural Rendering~\cite{DeferredNeuralRendering} takes a different step, by combining traditional graphics with learnable neural textures, embedded on the 3D facial mesh. Apart from the aforementioned person-specific approaches, there is a plethora of person-generic methods, which require only a few frames of the target identity \cite{wiles2018x2face,siarohin2019animating,siarohin2019first,averbuch2017bringing,geng2018warp,zakharov2019few,zakharov2020fast,ha2020marionette, doukas2021headgan}. 

\noindent{\textbf{Neural Representations for Scenes and Faces}.} With the introduction of NeRFs~\cite{mildenhall2020nerf}, much research has been focused on neural scene representations \cite{martin2021nerf, pumarola2021d}, with various attempts to model human faces. More specifically, Nerfies~\cite{park2021nerfies} and HyperNeRF~\cite{park2021hypernerf} have shown incredible results for reconstructing non-rigid scenes of moving heads. Despite their impressive generative ability, such systems are not able to control neither head poses nor facial movements. In a different direction, NerFACE~\cite{Gafni_2021_CVPR} proposes to control a dynamic NeRF with the assistance of expression blendshapes. In a similar line of work, AD-NeRF~\cite{guo2021ad} proposes an audio-driven NeRF-like model, based on acoustic features extracted with DeepSpeech \cite{hannun2014deep, amodei2016deep}.

\section{Method}
 
In this section, we first describe our baseline approach to the problem of full-head reenactment (Sec.~\ref{sec:2D_coordinate_based_MLP}), that is a 2D coordinate-based MLP with controllable dynamics. Then, we couple the MLP with a CNN-based decoder to build our full model for video portrait synthesis (Sec.~\ref{sec:dynamic_neural_portraits_method}), namely \textit{Dynamic Neural Portraits}. Finally, we show an extension of our system that supports audio-driven synthesis (Sec.~\ref{sec:audio}).

\subsection{2D MLP with Controllable Dynamics}
\label{sec:2D_coordinate_based_MLP}

Let $\mathbf{I}$ be an image of a fixed resolution $H \times W$. We can represent $\mathbf{I}$ with a neural network by training a fully-connected MLP to reconstruct the image from its 2D coordinates \cite{dupont2021coin}. For the synthesis of each pixel, we pass its coordinates $\mathbf{x}=(x,y)$ through the MLP network, which returns the colour of the pixel $\mathbf{c}$. We optimise the network by penalising the distance between the predicted and true colour. In order to compute the entire image, the MLP is evaluated at each position $(x, y)$ of the coordinate grid.

The model described above is quite limited, as it learns only to reconstruct a single static image from its pixel coordinates. We extend the 2D coordinate-based MLP for handling temporally varying data, such as video portraits of human faces. Here, we focus on RGB videos captured by a monocular and stationary camera. A time-varying representation can be obtained by conditioning the MLP network on facial information that changes between frames. Let $\mathbf{I}_{1:T}$ be a sequence of frames and $\mathbf{p}_{1:T}$, $\mathbf{e}_{1:T}$ be the corresponding head pose and facial expression parameters respectively, which have been recovered with a face tracking system. An intuitive solution for modeling the video portrait with a neural network would be to estimate the pixel's RGB value from the $i$-th video frame with the MLP, using the pixel's coordinates $\mathbf{x}$ and tracked parameters $\mathbf{p}_i, \mathbf{e}_i$, as
\begin{equation}
     \mathbf{c} = C(\mathbf{x}, \mathbf{p}_i, \mathbf{e}_i).
\end{equation}

Here, $\mathbf{c}$ is the colour at 2D location $\mathbf{x} \in [0, 1]^2$, $\mathbf{p}_i \in \mathbb{R}^6$ is the head pose that is given by rotation (Euler angles) and translation parameters, and $\mathbf{e}_i \in \mathbb{R}^{n_{exp}}$ are the expression parameters, which correspond to non-rigid facial deformations. Please note that the pose parameters $\mathbf{p}_i$ describe the rigid motions of the face with respect to the camera and do not include information regarding torso movements. The position of the camera remains fixed throughout the frames.

\subsection{Dynamic Neural Portraits}
\label{sec:dynamic_neural_portraits_method}

\begin{figure*}[t!]
\centering
\vspace{-1.25em}
\includegraphics[scale=0.363]{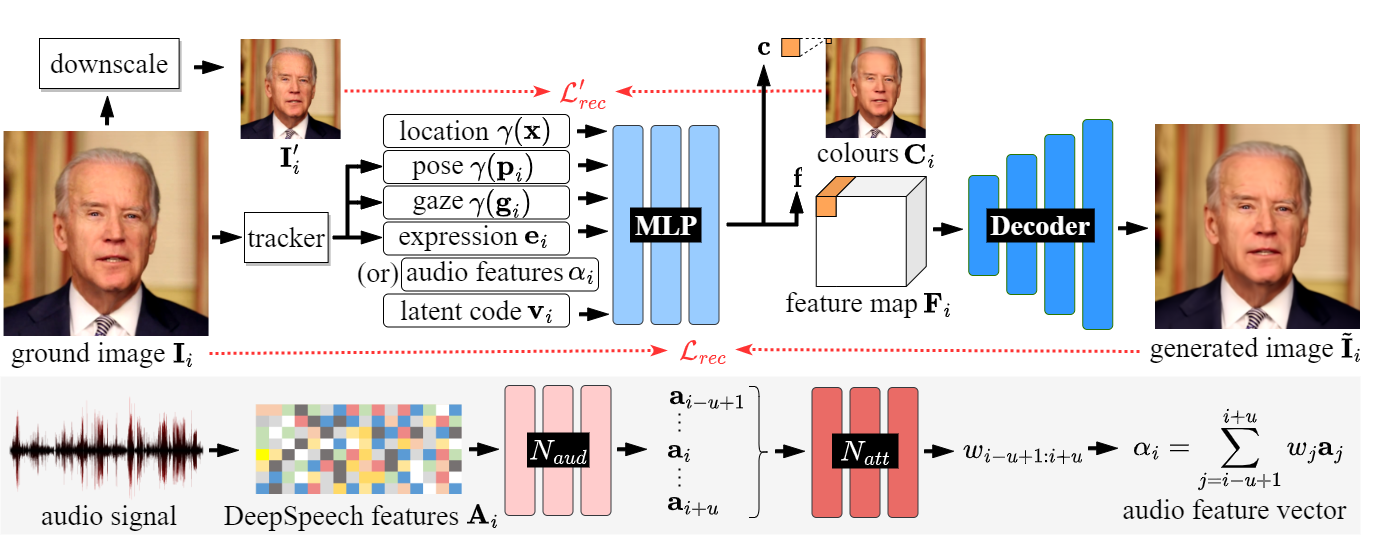}
\caption{An overview of \textit{Dynamic Neural Portraits} framework during training. As opposed to previous full-head reenactment methods that rely on image-to-image translation networks, our model is made up of an MLP encoder and a CNN decoder. Instead of facial expression parameters, we can drive synthesis using audio features, recovered from acoustic signals.}
\label{fig:NeuralPortraits}
\vspace{-1.25em}
\end{figure*}

Even though the 2D coordinate-based MLP with controllable dynamics is a straightforward approach for modeling video portraits with a single MLP network, in practice we found that the photo-realism of generated samples and the rendering speed significantly degrade as we increase the resolution of videos. As shown in our experiments, we obtain superior results in terms of visual quality by combing the MLP with a convolutional decoder network. Following the paradigm of recent 3D aware GANs \cite{niemeyer2021giraffe, gu2021stylenerf, zhou2021cips, chan2022efficient}, instead of predicting RGB colour values we propose an MLP that maps its input to a visual feature vector $\mathbf{f} \in \mathbb{R}^{n_f}$. Given the 2D spatial location $\mathbf{x}$ along with pose $\mathbf{p}_i$, expression $\mathbf{e}_i$ and gaze information $\mathbf{g}_i$, our MLP now predicts a feature vector
\begin{equation}
\mathbf{f} = F(\gamma(\mathbf{x}), \gamma(\mathbf{p}_i), \gamma(\mathbf{g}_i), \mathbf{e}_i, \mathbf{v}_i).
\end{equation}
We observed that we acquire more accurate eye movements by adding eye gaze angles $\mathbf{g}_i \in \mathbb{R}^2$ as an extra input to our MLP network. Moreover, we noticed that introducing per-image learnable latent variables $\mathbf{v}_i$ as inputs, a technique which has been previously adopted in NeRF methods \cite{martin2021nerf,park2021nerfies,park2021hypernerf,Gafni_2021_CVPR}, improves the stability of our network, since it enables the MLP to learn variations among frames that are not modeled by the pose and expression parameters (e.g. torso movements, illumination changes, small background motions). Furthermore, following established practices from NeRF-based systems and CIPS~\cite{anokhin2021image}, we adopt positional encoding for the MLP inputs, and more specifically on the position $\mathbf{x}$, pose $\mathbf{p}_i$ and gaze $\mathbf{g}_i$ vectors, which are low-dimensional. The standard encoding function
\begin{align}
\begin{split}
    \gamma(x) = [x, & \sin(2 \pi x), \cos(2 \pi x), \dots, \\
    & \sin(2^{N-1} \pi x), \cos(2^{N-1} \pi x)]^{\top}
\end{split}
\end{align}
proposed in \cite{mildenhall2020nerf} is applied to all values of $\mathbf{x}$, $\mathbf{p}_i$ and $\mathbf{g}_i$, mapping each number $x \in \mathbb{R}$ to a higher dimension embedding $\gamma(x) \in \mathbb{R}^{2 \cdot N + 1}$. Replacing the original inputs with their embeddings allows our model to achieve high frequency details in the generated images.

The MLP network described above learns to estimate a feature vector $\mathbf{f}$ for each spatial location of the image plane independently, while considering head pose, facial expression and eye gaze information. In order to render frame $i$, we first evaluate the MLP network at each spatial position $\mathbf{x} \in \mathbf{X}$ of the coordinate grid that corresponds to resolution $H_f \times W_f$, while keeping all other inputs fixed. After that, we accumulate the resulting features in a visual feature map $\mathbf{F}_i \in \mathbb{R}^{H_f \times W_f \times n_f}$. Then, we employ a decoding network $D$, which receives the feature map and performs up-sampling, in order to synthesise the output frame $\tilde{\mathbf{I}}_i = D(\mathbf{F}_i)$ at the target resolution $H \times W$. An overview of our proposed framework is shown in Fig.~\ref{fig:NeuralPortraits}. More details on the architecture of networks are available in the supplementary material.

\noindent{\textbf{Objective Function and Optimisation}.} We train our proposed model on the task of reconstruction. Given the generated image $\tilde{\mathbf{I}}_i$ and the corresponding ground truth frame $\mathbf{I}_i$, we define the reconstruction loss as the L2-distance between the predicted and true image. We experimented with L1, perceptual \cite{wang2018high} and adversarial \cite{ganGoodfellow} losses, but all of them produced substantially inferior results. However, we found that we can improve our method's performance by adding an extra output layer to the MLP, which predicts the colour $\mathbf{c} \in \mathbb{R}^3$ side-by-side with visual features $\mathbf{f}$, and minimise the distance of the predicted colour from the true one. To that end, we accumulate the colour outputs for all 2D spatial locations in $\mathbf{C}_i \in \mathbb{R}^{H_f \times W_f \times 3}$, and penalise the L2-distance from the ground truth image $\mathbf{I}_i'$ down-scaled to resolution $H_f \times W_f$, in order to match the resolution of $\mathbf{C}_i$. The overall loss term for frame $i$ is given as
\begin{equation}
    \mathcal{L} = \mathcal{L}_{rec} + \mathcal{L}_{rec}' = || \tilde{\mathbf{I}}_i - \mathbf{I}_i ||_2^2 + || \mathbf{C}_i - \mathbf{I}_i' ||_2^2
\end{equation}
We optimise the MLP network and CNN-based decoder jointly, in an end-to-end fashion.

\subsection{Audio-driven Portrait Synthesis}
\label{sec:audio}

Our choice to inject the driving signals (i.e. pose, expression and gaze parameters) into our system through an MLP network enables to easily adapt our method to other driving modalities, such as acoustic signals. Unlike previous works \cite{guo2021ad,Gafni_2021_CVPR,thies2020nvp,kim2018deep}, which focus exclusively either on expression blendshapes or audio data as driving signals, we show that the same architecture can be used effectively for both modalities. More specifically, we can replace the expression blendshapes with an audio feature vector $\boldsymbol{\alpha}_i$ as input to the MLP network, in case the audio stream is available. We extract high-level features from acoustic signals with the widely-adopted DeepSpeech model \cite{hannun2014deep, amodei2016deep}. As a first step, we assign a 29-dimensional vector estimated by DeepSpeech to every video frame. Then we create a window of vectors with size $w=16$ around each frame, taken from its neighboring (past and future) time steps. In this way each frame $i$ is coupled with a DeepSpeech feature $\mathbf{A}_i \in \mathbb{R}^{16 \times 27}$. We follow a similar approach with AD-NeRF \cite{guo2021ad} and utilise a 1D-convolutional network $N_{aud}$ that learns to compute per-frame latent codes $\mathbf{a}_i \in \mathbb{R}^{n_a}$ from $\mathbf{A}_i$. Next, we employ a self-attention network $N_{att}$, as proposed in \cite{thies2020nvp} and \cite{guo2021ad}, which operates as a temporal filter on subsequent audio codes $\mathbf{a}_{i-u+1:i+u}$ and mixes them up with the assistance of predicted attention weights $w_{i-u+1:i+u}$, to form the final audio feature vector $\boldsymbol{\alpha}_i = \sum_{j=i-u+1}^{i+u} w_j \mathbf{a}_j$. We set $u=4$, which results in a window of $2 u = 8$ time steps. 

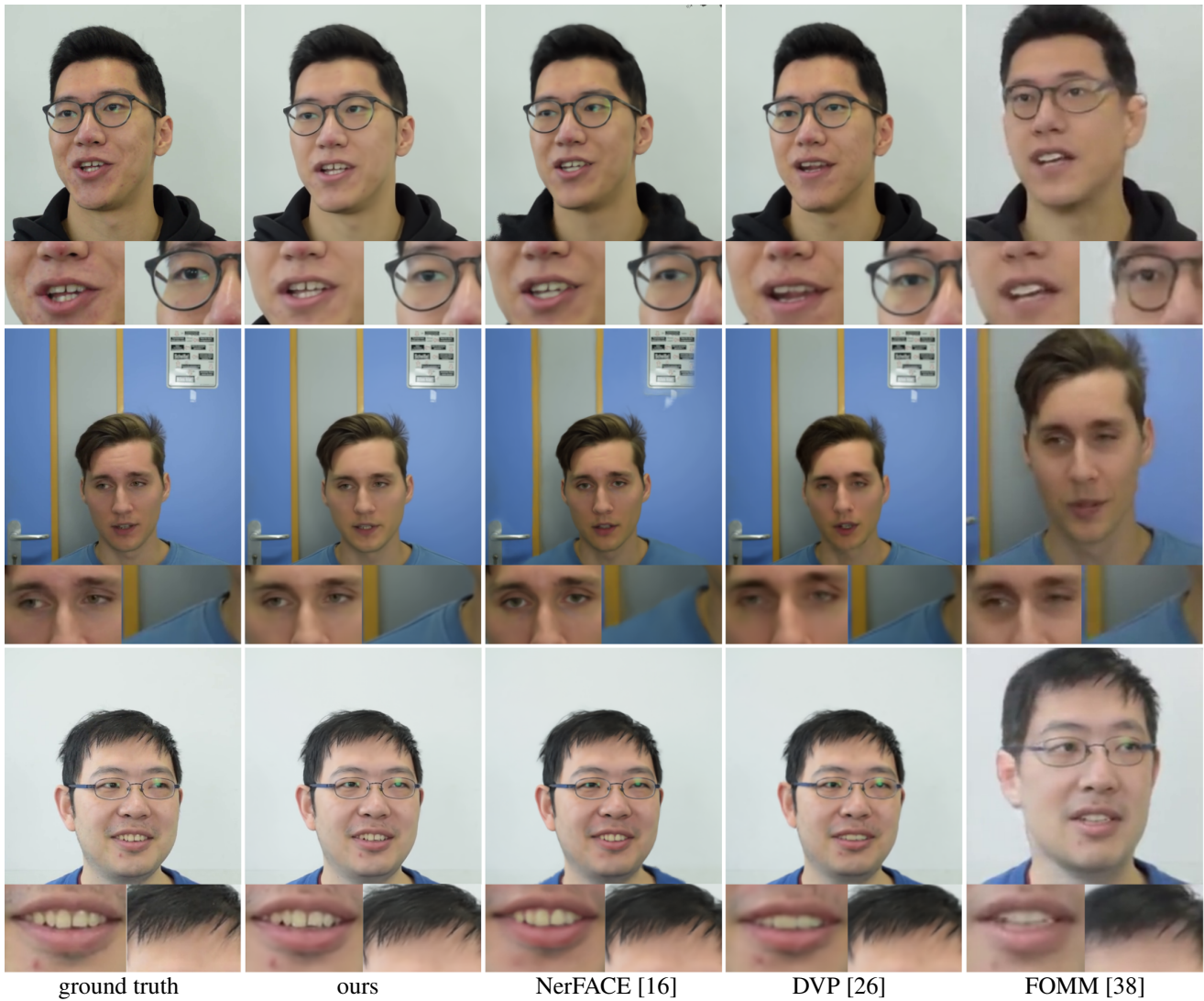
\begin{figure*}[b!]
\centering
\begin{picture}(100,405)
    \put(-197,10){\includegraphics[scale=0.224]{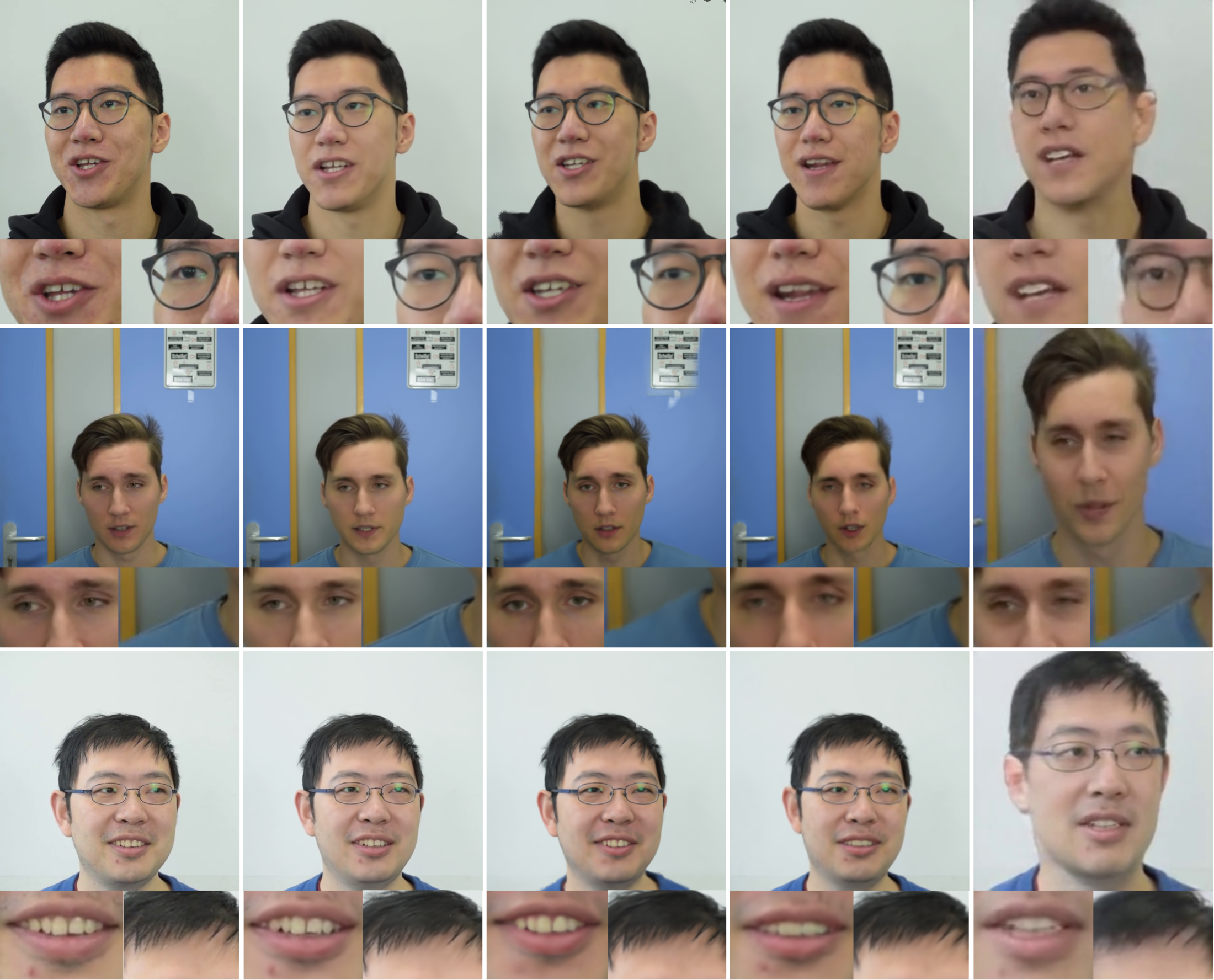}}
    \put(-175,1){ground truth}
    \put(-60,1){ours}
    \put(22,1){NerFACE~\cite{Gafni_2021_CVPR}}
    \put(128,1){DVP~\cite{kim2018deep}}
    \put(224,1){FOMM~\cite{siarohin2019first}}
\end{picture}
\caption{Visual comparison with baselines on the task of reconstruction. Our method consistently generates more realistic samples with finer details than its counterparts. For the evaluation of FOMM~\cite{siarohin2019first}, images had to be further cropped. Please zoom in for details.}
\label{fig:reconstruction_DNP}
\end{figure*}

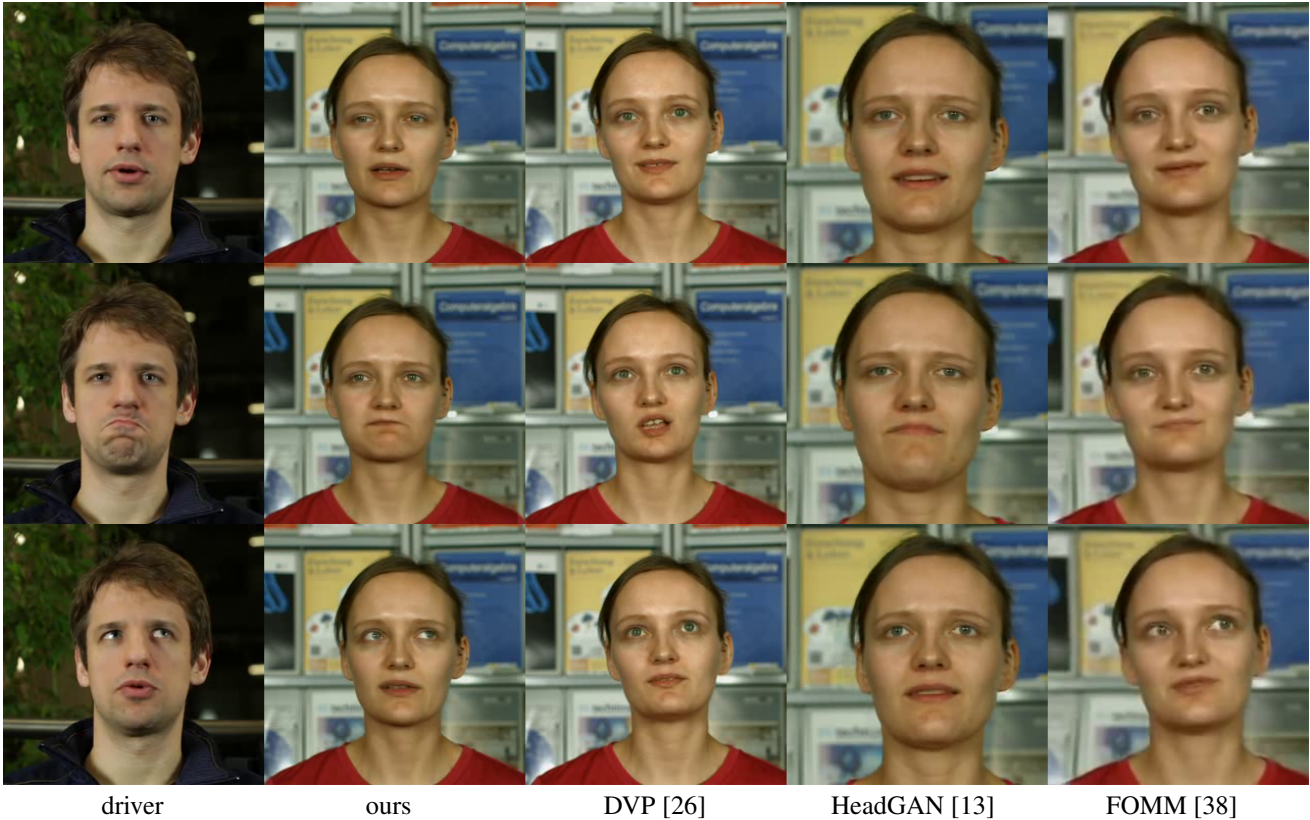
\begin{figure*}[b!]
\centering
\begin{picture}(100,302)
    \put(-197,12){\includegraphics[scale=0.384]{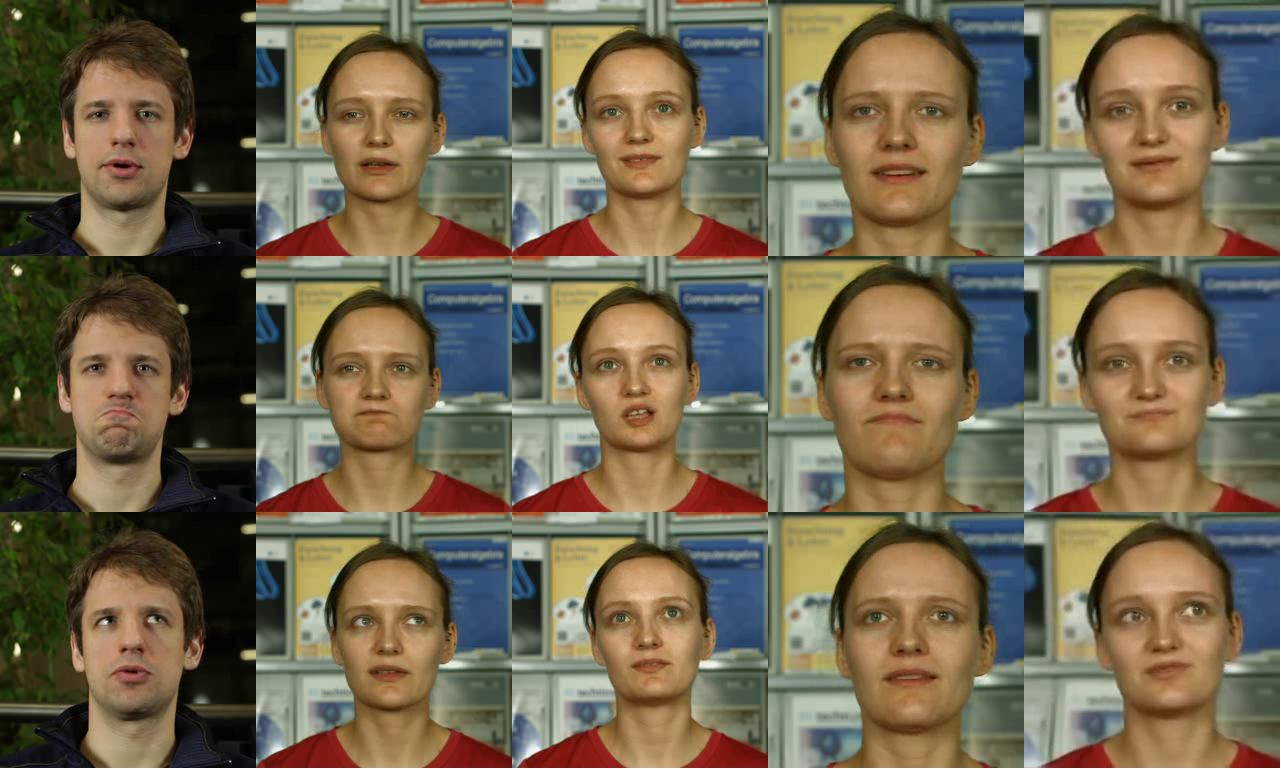}}
    \put(-160,1){driver}
    \put(-60,1){ours}
    \put(30,1){DVP~\cite{kim2018deep}}
    \put(116,1){HeadGAN~\cite{doukas2021headgan}}
    \put(220,1){FOMM~\cite{siarohin2019first}}
\end{picture}
\caption{Visual comparison with baselines on reenactment. Our approach transfers pose, expression and gaze more reliably than both person-specific DVP \cite{kim2018deep} and person-generic methods HeadGAN \cite{doukas2021headgan} and FOMM \cite{siarohin2019first}. Please note that for the correct evaluation of HeadGAN and FOMM, images had to be cropped closer to the face.}
\label{fig:reenactment_DNP}
\vspace{-0.5em}
\end{figure*}

\section{Experiments}
\label{sec:dynamic_neural_portraits_experiments}

\noindent \textbf{Dataset}. Our networks are optimised on monocular RGB videos, of various resolutions: $256^2$, $512^2$ and $1024^2$. We train a new model for each video portrait (different individual). As a pre-processing step, we crop frames around the target face and compute pose, expression and gaze parameters. We experiment with videos from 5K to 20K frames. In the supplementary material we provide more details of the adopted video database and face tracking systems.

\setlength{\tabcolsep}{4pt}
\begin{table}[t!]
\begin{center}
\small
\begin{tabular}{cccccc}
\toprule
Method & Portrait & L1 ($\downarrow$) & LPIPS ($\downarrow$) & FID ($\downarrow$) & FVD ($\downarrow$) \\
\midrule
\multirow{3}{*}{FOMM} & ID. 1 & 8.79 & 0.095 & 25.48 & 331.0 \\
& ID. 2 & 7.60 & 0.084 & 37.24 & 338.3 \\
& ID. 3 & 9.48 & 0.130 & 24.72 & 254.10 \\
\hline
\multirow{3}{*}{DVP} & ID. 1 & 6.95 & 0.152 & 49.35 & 195.4 \\
& ID. 2 & 9.08 & 0.079 & 37.58 & 464.3 \\
& ID. 3 & 8.01 & 0.123 & 51.30 & 196.7 \\
\hline
\multirow{3}{*}{NerFACE} & ID. 1 & \textbf{6.19} & 0.136 & 74.22 & 278.3 \\
& ID. 2 & 10.98 & 0.143 & 74.02 & 357.7 \\
& ID. 3 & 6.28 & 0.067 & 34.64 & 81.17 \\
\hline
\multirow{3}{*}{Ours} & ID. 1 & 6.45 & \textbf{0.094} & \textbf{23.76} & \textbf{169.8} \\
& ID. 2 & \textbf{5.21} & \textbf{0.071} & \textbf{24.60} & \textbf{222.1} \\
& ID. 3 & \textbf{5.15} & \textbf{0.051} & \textbf{23.94} & \textbf{78.06} \\
\bottomrule
\end{tabular}
\end{center}
\caption{Numerical comparison with FOMM \cite{siarohin2019first}, DVP \cite{kim2018deep} and NerFACE \cite{Gafni_2021_CVPR} for three different portraits on the task of reconstruction (self-reenactment).}
\label{table:reconstruction}
\vspace{-1em}
\end{table}

\subsection{Comparison with State-of-the-Art}

\noindent \textbf{Reconstruction.} First, we carry out a comparison with Deep Video Portraits (DVP)~\cite{kim2018deep} and NerFACE \cite{Gafni_2021_CVPR}, the best-performing person-specific reenactment methods, as well as First Order Motion Model (FOMM)~\cite{siarohin2019first}, a representative of person-generic models. We evaluate the generative performance of methods on the task of reconstruction, also know as self-reenactment. We assess the fidelity of reconstruction quantitatively, with the assistance of L1-distance between generated and ground truth test frames, as well as with Learned Perceptual Image Patch Similarity (LPIPS)~\cite{zhang2018unreasonable}, which tests the perceptual similarity between images. Moreover, we determine photo-realism with Fréchet Inception Distance (FID)~\cite{heusel2017gans} and Fréchet Video Distance (FVD)~\cite{unterthiner2018towards} metrics, which appear to correlate well with human perception. The results displayed in Table~\ref{table:reconstruction} indicate that our method outperforms all baselines, for three different video portraits. Please note that the reported numbers refer to the average scores, computed across all test frames for each video portrait. An exception to that are FID~\cite{heusel2017gans} and FVD~\cite{unterthiner2018towards} metrics, as they are computed on pairs of videos. All scores suggest that our approach produces superior samples in terms of visual quality and photo-realism. This observation is confirmed visually in Fig.~\ref{fig:reconstruction_DNP}. As can be seen, our method generates more crispy images with finer details and more consistent eye gaze. For a better visualisation of our results, we urge the reader to refer to our supplementary video.

\begin{table}[h!]
\begin{center}
\small
\begin{tabular}{ccccc}
\toprule
\multirow{2}{*}{Method} & \multirow{2}{*}{CSIM ($\uparrow$)} & Expression & Pose & Gaze  \\
& & Dist. ($\downarrow$) & Dist. ($\downarrow$) & Dist. ($\downarrow$) \\
\midrule
FOMM & 0.840 & 13.28 & 6.31$^\circ$ & 9.42$^\circ$ \\
HeadGAN & 0.755 & 14.17 & 3.26$^\circ$ & 6.35$^\circ$ \\
DVP & 0.861 & 11.95 & 4.93$^\circ$ & 8.37$^\circ$ \\
Ours & \textbf{0.885} & \textbf{10.69} & \textbf{2.57$^\circ$} & \textbf{4.74$^\circ$} \\
\bottomrule
\end{tabular}
\end{center}
\caption{Numerical comparison with FOMM~\cite{siarohin2019first}, HeadGAN~\cite{doukas2021headgan} and DVP~\cite{kim2018deep} on the task of cross-identity reenactment.}
\label{table:reenactment}
\vspace{-1em}
\end{table}

\noindent \textbf{Reenactment}. We further validate our model's performance on the task of cross-identity motion transfer (reenactment) in a quantitative way. This task involves passing on the head pose, facial expression and eye gaze from a driving actor to another target person, while preserving the identity of the later, as the two subjects are now different. We compare our approach with DVP \cite{kim2018deep}, FOMM \cite{siarohin2019first} and HeadGAN \cite{doukas2021headgan}. To that end, we use the source and target video portraits provided by the authors of DVP. For the numerical comparison, we measure the target's identity preservation with Cosine Similarity (CSIM), based on identity embeddings extracted with the assistance of ArcFace~\cite{deng2019arcface}. Moreover, we use DECA \cite{DECA} to regress pose and expression parameters, both from the driving and generated frames. Then, we compute the L1-distance between expression parameters, as well as the head rotation distance in degrees. Finally, we use the gaze estimator from \cite{chen2018appearance} to regress gaze vectors, and calculate their angular distance. In Table~\ref{table:reenactment}, we show that our method achieves better scores than all three baselines, across all metrics related to successful reenactment. In Fig.~\ref{fig:reenactment_DNP}, we showcase examples of frames where our method transfers pose, expression and gaze more accurately than DVP~\cite{kim2018deep}, HeadGAN~\cite{doukas2021headgan} and FOMM~\cite{siarohin2019first}.

\begin{figure}[t!]
\centering
\begin{picture}(100,140)
    \put(-69,10){\includegraphics[scale=0.251]{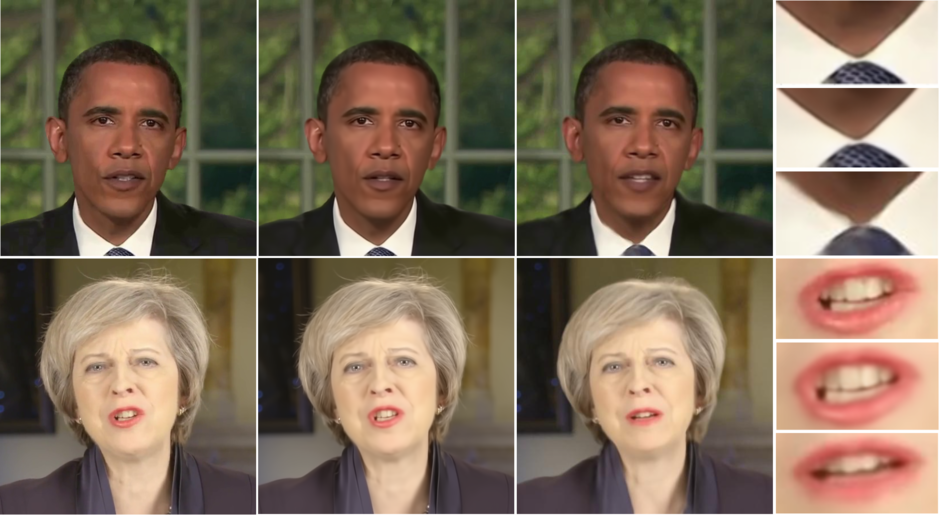}}
    \put(-60,1){ground truth}
    \put(18,1){ours}
    \put(64,1){AD-NeRF \cite{guo2021ad}}
\end{picture}
\caption{Visual comparison with AD-NeRF \cite{guo2021ad} on audio-driven reconstruction. We generate more accurate lips motions and higher quality details compared to AD-NeRF. Please zoom in for details.}
\label{fig:audio-reconstruction}
\vspace{-1em}
\end{figure}
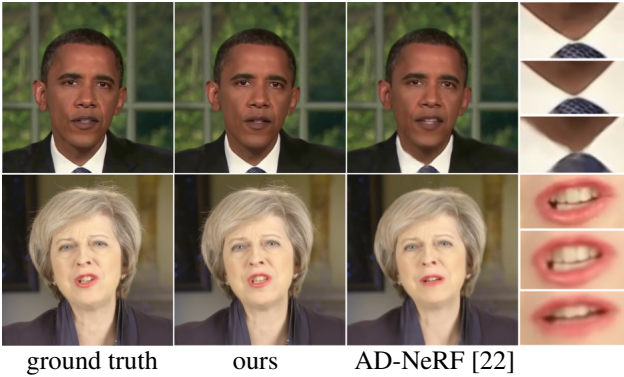

\noindent \textbf{Audio-driven reconstruction}. Except for conditioning our MLP network on expression blendshapes, we demonstrate the generative performance of our system when driven by acoustic signals. For that, we conducted a side-by-side comparison with AD-NeRF \cite{guo2021ad}, the state-of-the-art model on this task. As shown in Table~\ref{table:audio}, our quantitative analysis reveals that our approach performs better than AD-NeRF, both in terms of reconstruction and image quality. Our findings can be observed also visually in Fig.~\ref{fig:audio-reconstruction}.

\begin{table}[t!]
\begin{center}
\small
\begin{tabular}{cccccc}
\toprule
Method & Portrait & L1 ($\downarrow$) & LPIPS ($\downarrow$) & FID ($\downarrow$) & FVD ($\downarrow$) \\
\midrule
\multirow{2}{*}{AD-NeRF} & Obama & 4.11 & 0.083 & 16.67 & 225.7 \\
& May & \textbf{5.13} & 0.143 & 47.31 & 272.2 \\
\hline
\multirow{2}{*}{Ours} & Obama & \textbf{4.04} & \textbf{0.054} & \textbf{9.12} & \textbf{110.3} \\
& May & 5.72 & \textbf{0.087} & \textbf{26.72} & \textbf{96.1} \\
\bottomrule
\end{tabular}
\end{center}
\caption{Numerical comparison with AD-NeRF \cite{guo2021ad} on audio-driven video reconstruction.}
\label{table:audio}
\end{table}

\noindent \textbf{Execution Speed}. Owing to its lightweight architecture, our method is able to render frames in a resolution up to $1024 \times 1024$, in 24 fps. For lower resolutions (e.g. $256 \times 256$ and $512 \times 512$) our pipeline operates in speeds faster than real time. In comparison to recent NeRF-based state-of-the-art methods \cite{Gafni_2021_CVPR, guo2021ad}, our system achieves a significant speed-up, generating images nearly 270 times faster, which makes it much more efficient for real-world applications. In Table~\ref{table:times}, we report the execution times of different methods, measured on NVIDIA's Tesla V100 PCIe 32 GB. For DVP*~\cite{kim2018deep} we use the numbers recorded by its authors.

\begin{table}[t!]
\begin{center}
\small
\begin{tabular}{cccc}
\toprule
\multirow{2}{*}{Method} & $256 \times 256$ & $512 \times 512$ & $1024 \times 1024$ \\
 & time (fps) & time (fps) & time (fps) \\
\midrule
AD-NeRF~\cite{guo2021ad} & - & 9630 (0.10) & - \\
NerFACE~\cite{Gafni_2021_CVPR} & - & 8465 (0.12) & - \\
DVP*~\cite{kim2018deep} & 65 (15.4) & 196 (5.1) & - \\
HeadGAN~\cite{doukas2021headgan} & 41 (24.5) & - & - \\
FOMM~\cite{siarohin2019first} & 21 (47.2) & - & - \\
Ours & \textbf{11} (\textbf{90.9}) & \textbf{31} (\textbf{32.3}) & \textbf{42} (\textbf{24.2}) \\
\bottomrule
\end{tabular}
\end{center}
\caption{Comparison of the execution time between our generative model and related methods. Time is reported in milliseconds (msec). Please note that all reported numbers refer to the forward pass time of models during inference, without considering data pre-processing.}
\label{table:times}
\vspace{-1em}
\end{table}

\subsection{Ablation study}

\begin{figure*}[t!]
\centering
\begin{picture}(100,130)
    \put(-198,10){\includegraphics[scale=0.238]{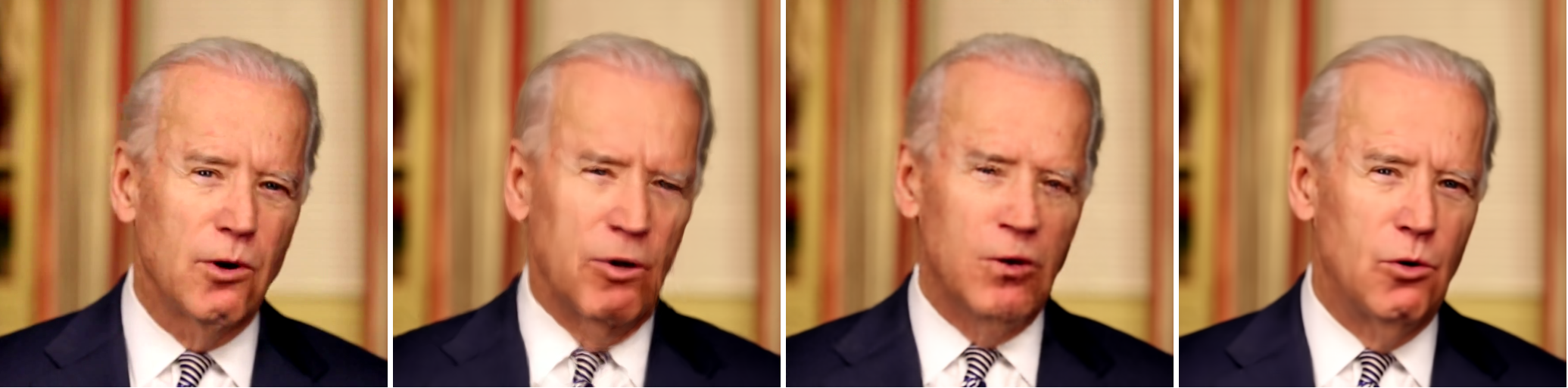}}
    \put(-162,1){ground truth}
    \put(-62,1){2D MLP contr. dyn. (A)}
    \put(64,1){CNN-based decoder (B)}
    \put(200,1){our full model (F)}
\end{picture}
\caption{Qualitative results of our ablation study. Our full model (F) improves the quality of frames synthesised by our baseline (A) (2D coordinate-based MLP with controllable dynamics) or the unaided CNN-based decoder (B) by a noteworthy margin. 
}
\label{fig:ablation_DNP}
\vspace{-0.3em}
\end{figure*}
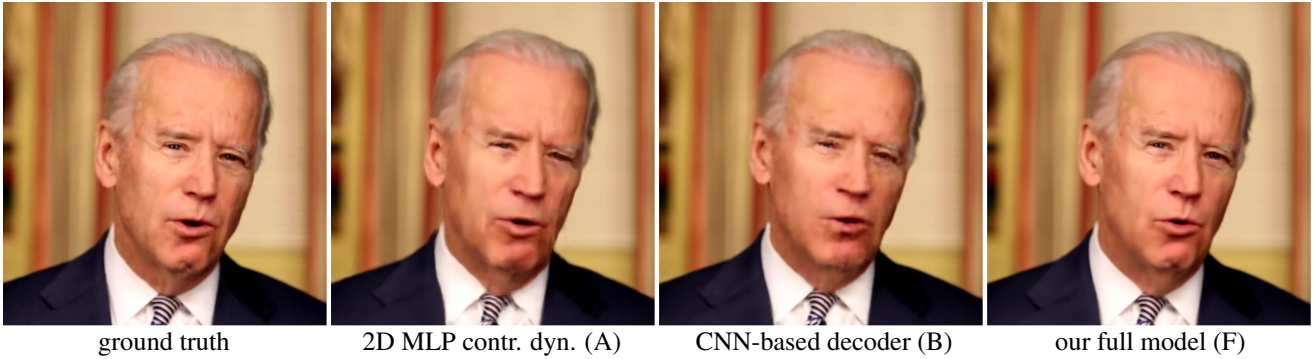

\begin{table*}[h!]
\begin{center}
\small
\begin{tabular}{clcccccc}
\toprule
\multicolumn{2}{c}{Variation} & Portrait & L1 ($\downarrow$) & LPIPS ($\downarrow$) & FID ($\downarrow$) & FVD ($\downarrow$) & Gaze Dist. ($\downarrow$) \\
\midrule
\multirow{2}{*}{(A):} & 2D coordinate-based MLP with & Biden & 5.33 & 0.072 & 10.69 & 197.2 & - \\
& controllable dynamics (baseline, Section \ref{sec:2D_coordinate_based_MLP}) & Obama & 3.74 & 0.077 & 14.33 & 132.9 & \\
\hline
\multirow{2}{*}{(B):} & \multirow{2}{*}{CNN-based decoder} & Biden & 6.67 & 0.089 & 11.51 & 263.8 & - \\
& & Obama & 4.27 & 0.068 & 12.78 & 127.4 & \\
\hline
\multirow{2}{*}{(C):} & \multirow{2}{*}{MLP + decoder} & Biden & 5.55 & 0.061 & 8.76 & 114.2 & - \\
& & Obama & 3.58 & 0.060 & 9.37 & 103.4 & \\
\hline
\multirow{2}{*}{(D):} & \multirow{2}{*}{(C) + latent codes} & Biden & 5.68 & 0.062 &\textbf{ 8.38} & \textbf{92.05} & - \\
& & Obama & 3.10 & 0.047 & 9.37 & 83.1 & \\
\hline
\multirow{2}{*}{(E):} & \multirow{2}{*}{(D) + $\mathcal{L}_{rec}'$} & Biden & 5.14 & 0.057 & 9.80 & 100.9 & 2.69 \\
& & Obama & \textbf{2.80} & 0.041 & 9.97 & 82.1 & 2.31 \\
\hline
\multirow{2}{*}{(F):} & (E) + gaze input & Biden & \textbf{5.03} & \textbf{0.054} & 9.42 & 107.2 & \textbf{1.86} \\
& (full model, Section \ref{sec:dynamic_neural_portraits_method}) & Obama & 2.86 & \textbf{0.041} & \textbf{9.33} & \textbf{79.3} & \textbf{2.18} \\
\bottomrule
\end{tabular}
\end{center}
\caption{Quantitative results of our ablation study on two separate video portraits (Biden and Obama), of resolution $512 \times 512$.}
\label{table:ablation_DNP}
\vspace{-1em}
\end{table*}

Next, we evaluate our design choices and validate the significance of different components that make up our model. We conduct quantitative and qualitative experiments for six variations of our system. To that end, we test all variations on the task of reconstruction, for two separate portraits (Biden and Obama). We start with a 2D coordinate-based MLP with controllable dynamics, described in Section \ref{sec:2D_coordinate_based_MLP}, which is used as the baseline model (A) of our ablation study. In order to demonstrate the advantages of our proposed architecture, we further experiment with a CNN-based decoding network only, without the MLP network (B). Then, we couple the MLP with the decoding network, which leads to variation (C). We form the next variations by first adding the learnable latent variables input (D), then including the reconstruction loss term $\mathcal{L}_{rec}'$ (E) and finally the gaze input (F), ending up with our full model as presented in Section \ref{sec:dynamic_neural_portraits_method}. Our numerical analysis presented in Table~\ref{table:ablation_DNP} reveals the importance of each component, especially the huge impact of the MLP network when coupled with a CNN-based decoding network. As can be observed, the latent variables increase FVD scores as they help to stabilise movements between frames. Furthermore, according to LPIPS, the $\mathcal{L}_{rec}'$ loss improves the perceptual similarity with ground truth data. Finally, the gaze input corrects eye motions, something that becomes more apparent when inspecting our supplementary video. In Fig.~\ref{fig:ablation_DNP}, we illustrate some examples of the visual differences between samples generated by variations (A), (B) and our full model (F). 

\section{Discussion}

\noindent \textbf{3D multi-view consistency}. Theoretically, our method is not 3D aware. Nonetheless, for the purposes of portrait reenactment this is not essential for achieving consistent results in various head poses. Given that training videos are captured by stationary cameras, they provide access to a single view of the scene. This makes 3D NeRF-based methods an over-parametrisation for such data, which end up violating 3D consistency for upper torso and producing severe shoulder trembling \cite{Gafni_2021_CVPR}. We argue that 3D modeling would require a video portrait captured by a moving camera from multiple viewpoints, as demonstrated most recently by RingNeRF~\cite{athar2022rignerf}. For videos captured by stationary cameras, our 2D model is able to generate frames with consistent appearance in diverse head poses and surpass 3D NeRF-based methods~\cite{Gafni_2021_CVPR, guo2021ad} in visual quality and inference speed.

\begin{figure}[t!]
\centering
\begin{picture}(100,90)
    \put(-67,10){\includegraphics[scale=0.6]{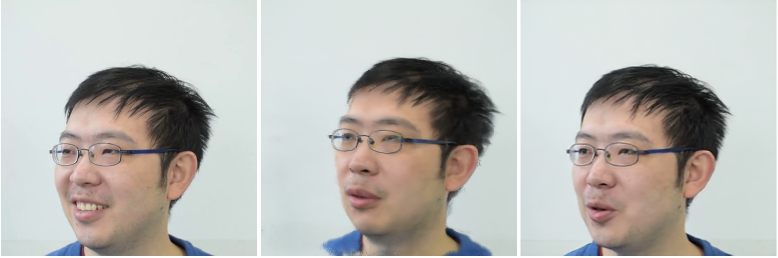}}
    \put(-67,2){max yaw (training)}
    \put(21,2){NerFACE~\cite{Gafni_2021_CVPR}}
    \put(118,2){ours}
\end{picture}
\caption{Evaluation of larger changes in head pose.}
\label{fig:largepose}
\vspace{-1.3em}
\end{figure}
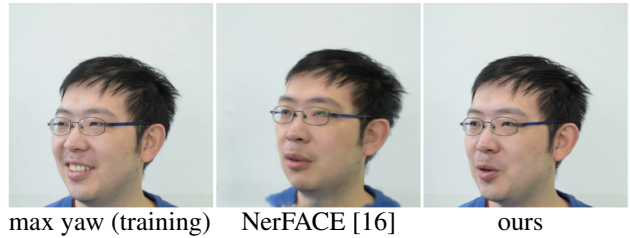

\noindent \textbf{Large changes in head pose}. The reenactment results we presented are mainly frontal. This is attributed to the training data, which consist mostly of frontal poses. Actually, both 2D and 3D-based methods are limited by the head pose variation that exists in the training video. We observed that all systems struggle to synthesise poses out of the training data span. In Fig. \ref{fig:largepose}, we illustrate the training frame with the most extreme head pose, in terms of yaw angle. Next to it, we show the most extreme pose synthesised by NerFACE~\cite{Gafni_2021_CVPR} and our method, without a substantial drop in quality. Our approach achieves higher photo-realism.

\noindent \textbf{Ethical considerations}. Generative models that offer explicit control on movements of faces could be potentially used for unethical purposes. For example, they could be used to synthesise videos of politicians acting in a provocative way. We would like to clarify that the intention of this paper is to make advancements realistic full-head reenactment for benevolent applications. We do not condone using our work to produce fake news or deceive the public.

\section{Conclusion}

We described \textit{Dynamic Neural Portraits}, a novel method for controllable video portrait synthesis. In contrast to previous attempts, our approach does not require renderings of 3D faces to drive synthesis and does not rely on GANs or NeRFs. Our experiments demonstrate the superiority of our model in terms of visual quality and run-time performance, compared to state-of-the-art video or audio-driven systems.

\vspace{0.22em}

\small
\noindent \textbf{Acknowledgement}. The work of Stefanos Zafeiriou was partially funded by the EPSRC Fellowship DEFORM: Large Scale Shape Analysis of Deformable Models of Humans (EP/S010203/1).

{\small
\bibliographystyle{ieee_fullname}
\bibliography{egpaper_arxiv}
}

\end{document}